\documentclass[10pt,twocolumn,letterpaper]{article}

\usepackage{iccv}
\usepackage{times}
\usepackage{epsfig}
\usepackage{graphicx}
\usepackage{amsmath}
\usepackage{amssymb}
\newcommand\blfootnote[1]{
    \begingroup
\renewcommand\thefootnote{}\footnote{#1}
    \addtocounter{footnote}{-1}
    \endgroup
}

\usepackage[breaklinks=true,bookmarks=false]{hyperref}

\usepackage{xspace}
\usepackage{color}
\usepackage[dvipsnames]{xcolor}
\usepackage{multirow}
\usepackage{adjustbox}
\usepackage{paralist} % compactitem, compactenum
\usepackage{enumitem} % adjust enum margin
\usepackage{booktabs} % Publication quality tables
\usepackage{array}
\usepackage{caption} % captionof
\usepackage{mathrsfs}
\usepackage{makecell}
\usepackage{xcolor}

\usepackage{comment}
\usepackage{multicol}
\usepackage{capt-of}
\usepackage{flushend}
\definecolor{gray}{rgb}{0.35,0.35,0.35}
\definecolor{MyBlue}{rgb}{0,0.2,0.8}
\definecolor{MyRed}{rgb}{0.8,0.2,0}
\definecolor{MyGreen}{rgb}{0.0,0.4,0.1}
\definecolor{MyGray}{rgb}{0.4,0.4,0.4}

\long\def\ignorethis#1{}

\usepackage[capitalize]{cleveref}
\crefname{section}{Sec.}{Secs.}
\Crefname{section}{Section}{Sections}
\Crefname{table}{Table}{Tables}
\crefname{table}{Tab.}{Tabs.}

\iccvfinalcopy % *** Uncomment this line for the final submission

\def\iccvPaperID{****} % *** Enter the ICCV Paper ID here

\ificcvfinal\fi

\begin{document}

%%%%%%%%% TITLE
\title{Implicit and Explicit Commonsense for Multi-sentence Video Captioning\vspace{-1em}}

\author{Shih-Han Chou$^{1,2}$, James J. Little$^{1}$, Leonid Sigal$^{1,2,3}$\\
$^{1}$Department of Computer Science, University of British Columbia\\
$^{2}$Vector Institute for AI \hspace{2mm}
$^{3}$Canada CIFAR AI Chair\\
\{shchou75, little, lsigal\}@cs.ubc.ca
}

\twocolumn[{%
\renewcommand\twocolumn[1][]{#1}%
\maketitle
\begin{center}
    \centering
    \vspace{-3em}
    \captionsetup{type=figure}
    \includegraphics[width=.75\textwidth]{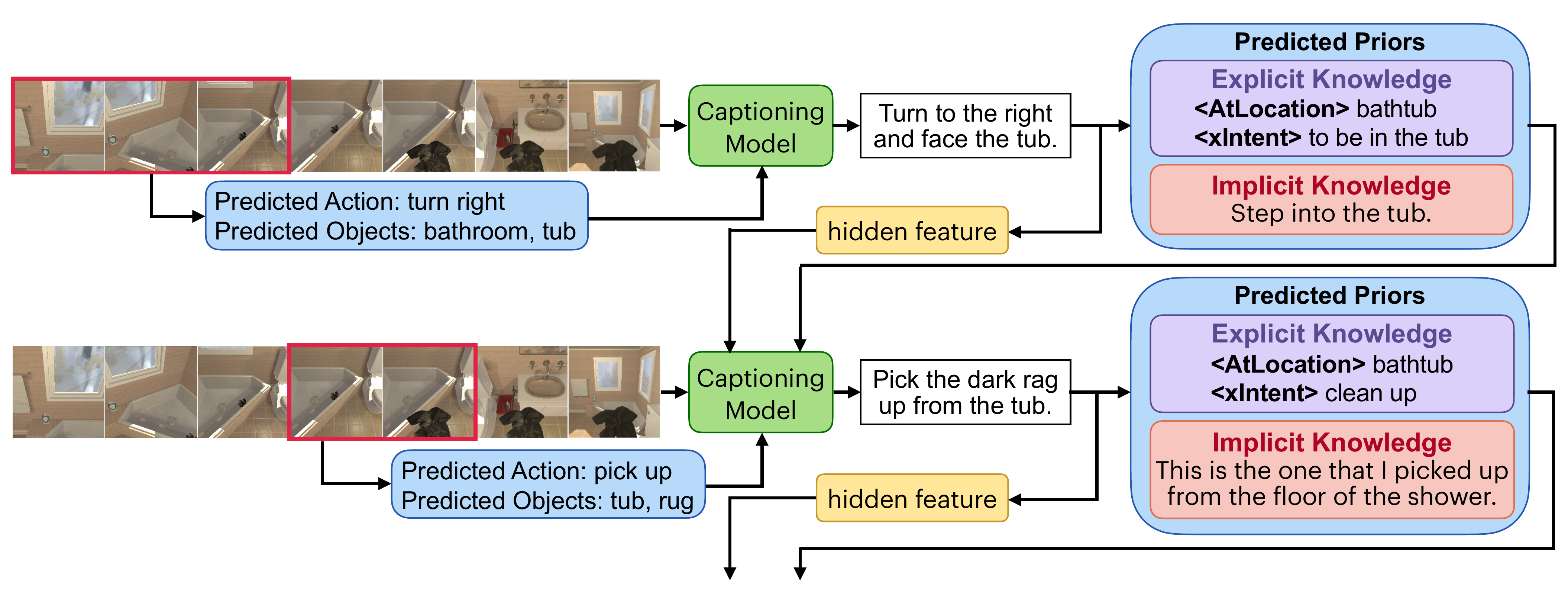}
    \vspace{-0.1in}
    \captionof{figure}{\textbf{Commonsense-enhanced Multi-sentence Video Captioning.} Illustration of the proposed multi-sentence video captioning that leverages implicit (sentence completion) and explicit (knowledge-base) forms of commonsense knowledge. For each video segment (snippet), we first predict the duration as well as the actions and objects. The captioning model takes these predictions along with inferred commonsense priors and hidden features, obtained from the previous snippet, to generate the caption for the current video segment.}\label{fig:frontfig}
\end{center}%
}]

\ificcvfinal\thispagestyle{empty}\fi

%%%%%%%%% ABSTRACT
\begin{abstract}
\vspace{-1em}
Existing dense or paragraph video captioning approaches rely on holistic representations of videos, possibly coupled with learned object/action representations, to condition hierarchical language decoders. 
However, they fundamentally lack the commonsense knowledge of the world required to reason about progression of events, causality, and even the function of certain objects within a scene. 
To address this limitation we propose a novel video captioning Transformer-based model, that takes into account both implicit (visuo-lingual and purely linguistic) and explicit (knowledge-base) commonsense knowledge. We show that these forms of knowledge, in isolation and in combination, enhance the quality of produced captions. 
Further, inspired by imitation learning, we propose a new task of instruction generation, where the goal is to produce a set of linguistic instructions from a video demonstration of its performance. We formalize the task using the ALFRED dataset~\cite{shridhar2020alfred} generated using an AI2-THOR environment.  
While instruction generation is conceptually similar to paragraph captioning, it differs in the fact that it exhibits stronger object persistence, as well as spatially-aware and causal sentence structure.
We show that our commonsense knowledge enhanced approach produces significant improvements on this task (up to 57\% in METEOR and 8.5\% in CIDEr), as well as the state-of-the-art result on more traditional video captioning in the ActivityNet Captions dataset~\cite{krishna2017dense}.$^*$\blfootnote{{*}The paper is under consideration at Computer Vision and Image Understanding Journal.}
\end{abstract}

%%%%%%%%% BODY TEXT
\vspace{-2em}
\section{Introduction}

Multi-modal visio-lingual models are becoming increasingly popular with tasks such as captioning \cite{anderson2018bottom, sharma2018conceptual, ravi2022vlc, seo2022end}, question answering \cite{anderson2018bottom, VQA, goyal2017making, Chadha2020iPerceive, chou2022semi, ben2017mutan}
and grounding \cite{xiao2017weakly, bajaj2019g3raphground, deng2018visual, rohrbach2016grounding}.
Among these, video tasks, such as video captioning, tend to be particularly challenging as they require an understanding of both visual content within frames and across frames to generate compelling and accurate descriptions. 
Early video captioning models employed global encoder-decoder architectures to caption short (snippet) videos \cite{Chadha2020iPerceive, seo2022end}.
More recent works focus on long untrimmed videos and the generation of lengthy and detailed descriptions. 
For example, the task of {\em dense video captioning} \cite{krishna2017dense} focuses on first identifying temporal segments of the video worth describing and then captioning those segments, each with a single sentence. 
% As a whole, 
The result is a longer multi-sentence story-like grounded description. 
A related task is {\em paragraph captioning}~\cite{song2021towards,liu-wan-2021-video}, where the output is similar, but the task does not require temporal localization in % the 
video. 
% itself. 
Significant progress has been made in generating multi-sentence descriptions for those video tasks. 
A core challenge is the context that needs to be propagated between sentences, which in past works has been often addressed with hierarchical decoders \cite{xiong2018move,xu2019joint} and memory-augmented networks \cite{lei2020mart}.
Flexible conditioning \cite{mehri2018middle} and diversity of generated captions \cite{song2021towards} have also been explored.  
However, all existing models, with exception of \cite{Chadha2020iPerceive}, learn directly from video-caption pairs and lack any notion of commonsense knowledge such as information on the function of objects, plausible cause and effect relations, and other forms of prior knowledge that one may imagine being very useful in story-telling. 
Consider a video depicting a person picking up a mop. A human can easily recognize a function of the mop -- to clean floors, and can imagine the events that will likely follow (\eg, person wetting a mop and then cleaning the floor). 
We posit that incorporating this type of commonsense knowledge into captioning would both improve the performance and result in multi-sentence captions that are more consistent and contain better discourse. 

Modeling of commonsense knowledge has a rich history in AI. 
Recent works can be categorized into two classes: {\em explicit} and {\em implicit}. 
Explicit models such as ConceptNet \cite{speer2017conceptnet}, ATOMIC \cite{sap2019atomic} and COMET \cite{Hwang2021COMETATOMIC2O} attempt to encode taxonomic, physical and social knowledge from human-labeled entity-relation tuples.  
Implicit models, such as GPT \cite{brown2020language} and variants, simply predict the next sentence succeeding the current sentence. It has been shown that such expressive language models can indeed encode aspects of commonsense knowledge including causality \cite{Chatpatanasiri2021_GPT3CommonSense,jiang2020tacl, trinh2018}. 

In this paper, we propose a new task of {\em instruction generation}, where, given a first-person video of an actor performing a task, the goal is to generate a set of (instructional) sentences that characterize the series of actions that need to be undertaken to perform it. 
Instruction generation is motivated by imitation learning, and, importantly, requires the generation of a series of sentences that are closely inter-related. 
Specifically, formalizing instructions requires (1) persistence (\eg, a picked up object should be utilized and named consistently in each instruction it is involved in); (2) spatial awareness, \eg, for navigation, which is a form of commonsense knowledge; and (3) causal and functional inferences (\eg, a {\em dirty rug} can be {\em washed} which will make the {\em rug} clean and likely require {\em drying} afterwords; see Figure~\ref{fig:frontfig}). 
In other words, instruction generation requires at least a certain level of commonsense knowledge. 

Further, we propose a video captioning model that incorporates both explicit (from COMET \cite{Hwang2021COMETATOMIC2O}) as well as implicit commonsense knowledge in the form of a language model (GPT3 \cite{brown2020language}) and visio-lingual zero-shot object knowledge (through CLIP \cite{radford2021learning}). 
Our video captioning model extracts a snippet of video, detects objects and actions in it, and uses encoding of the snippet along with object/actions, CLIP detected objects, and commonsense inferences (that are prompted with the previous sentence) to produce a new sentence. This procedure is repeated recursively until the desired number of sentences are decoded and/or the end of the paragraph is reached. 
We explore both Transformer-based (trained directly for the task) and LLM-based foundational language (fine-tuned) decoders to show the generality of the proposed approach.
We also evaluate our approach on the traditional dense video captioning task within the ActivityNet Captions~\cite{krishna2017dense} dataset. Instruction generation and dense video captioning are related in the sense that they share similar inputs (video) and outputs (multiple sentences) but the differences are mainly in the nature of those inputs/outputs. We show that our approach that incorporates commonsense is able to improve performance for both tasks.

\vspace{0.1in}
\noindent
{\bf Contributions.}
Our contributions are three-fold:
(1) We propose a new task of video-based instruction generation, which we believe requires a certain amount of commonsense knowledge;
(2) We propose a new video multi-sentence captioning model that effectively leverages implicit and explicit commonsense knowledge (prompted with prior sentences) to improve current sentence prediction. It shows that commonsense inferences that go beyond what is \textit{seen} in the video are helpful in generating better and more precise captions/instructions.
(3) We explicitly analyze contributions these sources or knowledge make towards improved caption quality and also illustrate our model's performance on more traditional dense video captioning tasks, showing state-of-the-art performance (without large pre-training).
Overall, our paper is one of the first to to explore the use of commonsense knowledge for video captioning. \footnote{We will release the code and all pre-trained models.}

\section{Related Works}
\noindent \textbf{Video Captioning.}
Video captioning was first tackled with rule-based models~\cite{das2013thousand, kojima2002natural}. These methods detect the actions and objects in the video frames and use templates to combine them into sentences. With the success of machine translating systems~\cite{DBLP:journals/corr/BahdanauCB14,rohrbach2013translating, sutskever2014sequence}, later works~\cite{venugopalan2015sequence,yao2015describing} moved away from rule-based methods and framed the captioning task as one of machine translation with encoder-decoder designs. 
Specifically, the encoder takes a set of video features and accumulates the hidden state which then is passed to a decoder for captioning. Early works extend the visual encoder from a $2$D CNN module to $3$D CNNs~\cite{carreira2017quo, xie2017rethinking}. Chen \etal~\cite{chen2019motion} use attention strategies to temporally aggregate the entire video for better-capturing motion dynamics. To alleviate the computational challenge of using expensive $3$D CNNs applied to dense frame inputs, Seo \etal~\cite{seo2022end} use a Transformer-based encoder applied to raw pixels, sampled at a coarse rate to better capture long context. 
To leverage commonsense, Chadha \etal~\cite{Chadha2020iPerceive} propose iPerceive which generates common-sense features by inferring causal relationships between events in videos. In contrast, we leverage commonsense as priors for the next sentence generation.
Mun \etal~\cite{mun2019streamlined} utilize the ``context'' derived from prior {\em inferred} captions to generate the next caption; our method, in contrast, capitalizes on the commonsense knowledge that cannot be directly acquired from the video data -- showcasing our model's ability to imagine alternative insights.

\vspace{0.1in}
\noindent \textbf{Video Understanding.}
A significant milestone in the domain of video understanding was reached through dense video captioning~\cite{krishna2017dense}, which is inspired by the dense image captioning task~\cite{johnson2016densecap}. 
The ActivityNet Captioning \cite{krishna2017dense} dataset is released as a benchmark for this task and has generated a flurry of research~\cite{Chadha2020iPerceive, iashin2020better, iashin2020multi, seo2022end, song2021towards, wang2018bidirectional, zhang2022unifying, zhou2018end}.
For example, Wang \etal~\cite{wang2018bidirectional} propose attentive fusion to differentiate captions from highly overlapped events. Zhou \etal~\cite{zhou2018end} use a Transformer-based model to alleviate the limitations of RNNs to model long-term dependencies in videos. Recently, Song \etal~\cite{song2021towards} model dense video captioning as a multi-sentence task by leveraging dynamic video memory and directly generating a paragraph. 
Consequently, we build on top of this approach and compare to it as a natural baseline for our method.

Another line of research in video understanding has focused on instructional videos~\cite{miech2019howto100m, shridhar2020alfred}. Instructional videos contain snippets with corresponding instructions. This type of video is different from the traditional video captioning datasets~\cite{krishna2017dense, wang2018video} which lack spatially grounded and action-oriented language descriptions. In our work, we propose the {\em instruction generation} task. Given a first-person instructional video with an actor performing a task, the goal is to generate a set of (instructional) sentences that characterize the series of actions required to perform it.

\vspace{0.02in}
\noindent \textbf{Commonsense Reasoning. }
Commonsense reasoning and language modeling are widely used in NLP tasks. ConceptNet~\cite{speer2017conceptnet} focuses on taxonomic and lexical knowledge (such as {\em Related-To}, {\em Synonym}, {\em Is-A}) and physical commonsense knowledge (such as {\em Made-Of}, {\em Part-Of}). It contains $3.4$M entity-relation tuples in English. 
The ATOMIC~\cite{sap2019atomic} knowledge graph contains $1.33$M tuples across $9$ relations covering social commonsense knowledge including {\em xIntent}, {\em xWant}, {\em oWant}, \etc). 
Nowadays, several works incorporate knowledge into downstream tasks, for example, encoding subgraphs of relevant knowledge~\cite{garderes2020conceptbert,lin2019kagnet} or pre-training on commonsense knowledge bases~\cite{porada-etal-2022-pre,zhou2020pre}. 
However, ConceptNet, ATOMIC and other {\em explicit} common reasoning methods that leverage annotated tuples, suffer from limited coverage
(\ie, unable to make inferences on unseen, during training, entities or relations). 
COMET~\cite{Hwang2021COMETATOMIC2O}, a Commonsense Transformer is designed to address these problems by fine-tuning pre-trained language models on KBs. COMET is able to generate inferences for commonsense relations for new inputs. Various approaches~\cite{chakrabarty-etal-2022-rocket,shwartz-etal-2020-unsupervised,ravi2022vlc,majumder-etal-2020-like,tian-etal-2021-hypogen-hyperbole} leverage COMET to generate knowledge for language tasks. In our work, we use COMET as the {\em explicit} commonsense prior in multi-sentence generation (\eg, of instructions).

Recently, NLP has had a series of breakthroughs with large-scale language modeling. In general, language modeling estimates the probability of the next word (token / symbol) in a sequence. Approaches are trained in a self-supervised manner from vast corpora of data.  
BERT~\cite{kenton2019bert} model is a Transfer-based encoder that significantly advanced language understanding~\cite{wang2019superglue}. 
For decoder-only models, the Generative Pretrained Transformer (GPT) models~\cite{radford2018improving,raffel2020exploring} set the state-of-the-art in language modeling performance. The most recent variant, GPT-3~\cite{brown2020language} is able to be adapted to various tasks, such as text generation, by using few-shot techniques. 
In our model, we take the generated text from GPT-3 as the {\em implicit} commonsense prior to guide the next sentence generation.

\section{Approach}

\begin{figure*}[!t]
    \centering
    \includegraphics[width=0.9\textwidth]{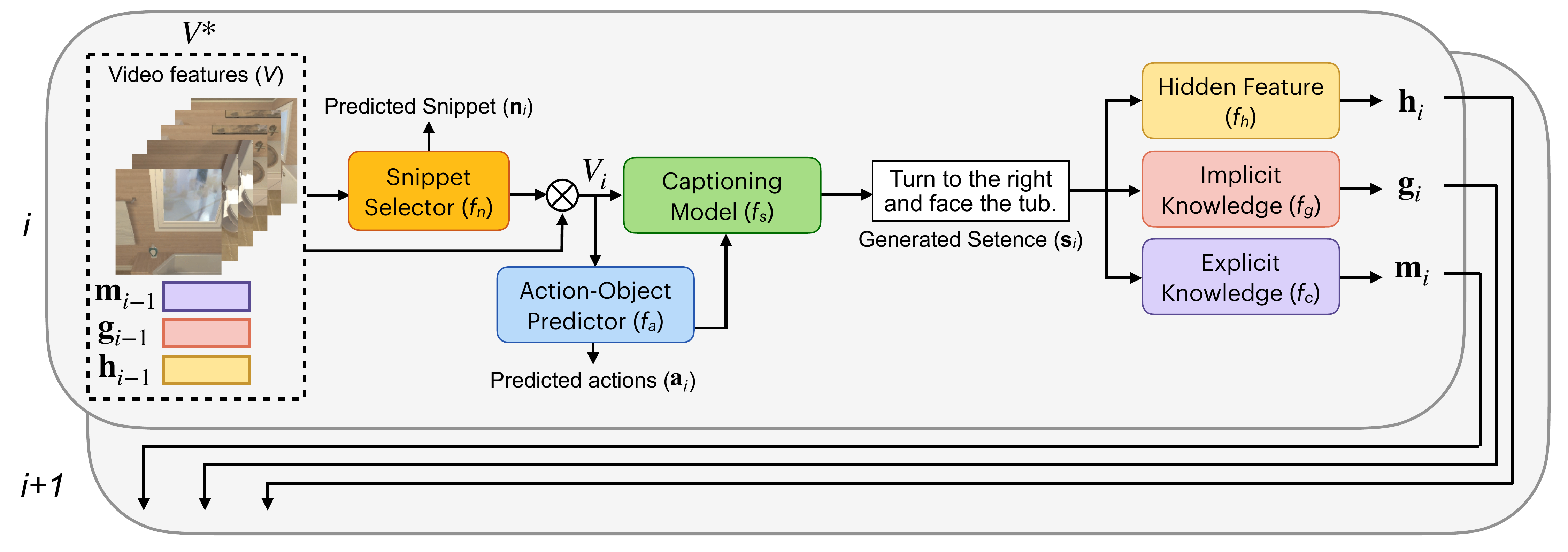}
    \caption{Model Overview of the Proposed Video Captioning Transformer-based Model. 
    The proposed model takes the video features, commonsense knowledge features, and hidden features as inputs for the snippet selector to predict the duration of the snippet. The video features are then multiplied with the predicted snippet to form the snippet features. The Action-Object predictor uses the snippet features, commonsense knowledge features, and hidden features to predict the actions and objects that appear in the snippet. In the captioning model, the concatenation of the snippet features, commonsense knowledge features, hidden features, and predicted actions-objects are fed into the Transformer encoder-decoder to generate the sentence which then passes through the implicit and explicit knowledge module to predict the commonsense knowledge for the next snippet.
    }
    \label{fig:model}
\end{figure*}

In this section, we introduce the proposed video captioning Transformer-based model. Figure~\ref{fig:model} illustrates the overall model pipeline. Given per-frame video features $\mathcal{V} \in \mathbb{R}^{T\times 4096}$, obtained from the concatenation of ResNet-200 \cite{he2016deep} features $\in \mathbb{R}^{T\times 2048}$, I3D RGB \cite{carreira2017quo} features $\in \mathbb{R}^{T\times 1024}$, and I3D Flow \cite{carreira2017quo} features $\in \mathbb{R}^{T\times 1024}$, as inputs, our goal is to generate the corresponding captions $\mathcal{S}=\{\mathbf{s}_1, \mathbf{s}_2, ..., \mathbf{s}_C\}$, where $T$ is the video length, and $C$ is the number of snippets (or video segments) chosen to be captioned\footnote{Both $T$ and $C$ would differ for each video, but we omit sample referencing to avoid notational clutter.}. In order to generate more contextual and coherent text, we integrate two kinds of commonsense knowledge, {\em explicit} and {\em implicit}, into our model (see Sec.~\ref{subsec:prior} for details). 

The overall model is described in Sec.~\ref{subsec:sentgen} and operates autoregressively in three steps.
First, in a given step $i$, video features, $\mathcal{V}$, and provided context are used by the  {\em snippet predictor} to soft select the next temporal segment $\mathbf{n}_i \in \mathbb{R}^T$ of video to caption.
Second, {\em Action-Object predictor} is used to detect objects and actions $\mathbf{a}_i$ within the pulled representation of the frames from the selected segment. In this process we leverage zero-shot object detection, through CLIP \cite{radford2021learning}, and supervised object/action classification when labels are available. 
Finally, the Transformer-based captioning model generates the new caption, $\mathbf{s}_i$ for the corresponding video segment. In doing so we leverage object and action labels $\mathbf{a}_i$ estimated earlier.
Importantly, in the last and also first step, where we select the next segment, for added semantic context,  we leverage {\em explicit} and {\em implicit} commonsense knowledge, $\mathbf{m}_i$ and $\mathbf{g}_i$, respectively, encoded by SBERT \cite{reimers-2019-sentence-bert} embeddings of the inferences and sentence completions generated by COMET~\cite{Hwang2021COMETATOMIC2O} and GPT3 \cite{brown2020language}.

Our proposed model is trained in a supervised manner from pairs of video and captions, with annotated video segments and corresponding aligned sentences. We discuss training and the learning objective in Section~\ref{subsec:loss}.

\subsection{Commonsense Knowledge Generation}\label{subsec:prior}

\noindent{\textbf{Explicit Commonsense Knowledge.}}\label{subsubsec:common}
In videos, the next snippet content is usually related to the current snippet. For example, if the current snippet is ``walk to the coffee maker on the right" and the following snippet is ``pick up the dirty mug from the coffee maker", we can observe that the agent needs to first walk to the coffee maker so that it can continue to the next action. 
% , ''pick up the dirty mug``. 
To model these relationships, we employ the most recent version of COMET~\cite{Hwang2021COMETATOMIC2O} initialized using BART~\cite{lewis-etal-2020-bart} in a zero-shot setting. The COMET is trained with $50$ relation types 
% including $\texttt{<AtLocation>, <isBefore>}$, \etc. 
from ConceptNet~\cite{speer2017conceptnet} and 
% $\texttt{<xIntent>, <xNeed>}$, etc. 
from ATOMIC~\cite{sap2019atomic}. We generate inferences based on the selected $12$ relation types, such as $\texttt{<AtLocation>, <xNeed>}$, that are most relevant and meaningful to our work and supported by COMET (The full list of the relations is in the supplementary material.). For each relation type, we take the top $1$ inference. % as it is the most related one to the next snippet. 
For example, take the input sentence $\mathbf{s}_{i-1}$=``walk to the coffee maker on the right" as an example; COMET can generate inferences such as \texttt{<AtLocation>} kitchen, \texttt{<xEffect>} gets coffee, \texttt{<IsAfter>} walk to kitchen, \etc. 
Finally, each inference is converted to a natural language form and concatenated to a large sentence\footnote{We also considered encoding each inference separately, but this resulted in high-dimensional context that was difficult to handle.} as ``At Location kitchen \texttt{<PAD>} ... Effect gets coffee \texttt{<PAD>} ..." with a \texttt{<PAD>} special token to separate relation types. 
To encode the generated commonsense for our model, we leverage the pretrained sentence Transformer (SBERT)~\cite{reimers-2019-sentence-bert}, resulting in commonsense feature vector, $\mathbf{m}_i \in \mathbb{R}^{384}$, for the current snippet. 

\vspace{0.07in}
\noindent{\textbf{Implicit Commonsense Knowledge.}}\label{subsubsec:gpt}
For the implicit commonsense knowledge, we observe that the next sentence is usually conditioned on the previous one in a paragraph. To leverage this phenomenon, we formulate it as a text completion problem. For a given text, the GPT-Neo model~\cite{black2021gpt} pretrained on The Pile dataset~\cite{gao2020pile} is used for the text completion task. The GPT-Neo model is a Transformer model designed to reproduce the GPT-3 architecture. We take the first generated sentence from the GPT-Neo model outputs as implicit commonsense knowledge. For example, for the input sentence, $\mathbf{s}_{i-1}$=``Pick the dark rag up from the tub.", we are able to get the next sentence as ``Pour the water from the bowl onto it." Similarly to the explicit commonsense, we leverage the pretrained SBERT model to encode the completed sentence into contextual commonsense representation for our model: $\mathbf{g}_i \in \mathbb{R}^{384}$.

\subsection{Video Captioning Transformer-based Model}\label{subsec:sentgen}

\noindent{\textbf{Snippet Selector. }}\label{subsubsec:snippets}
Considering that there are several snippets in videos that correspond to different sentences, we propose a snippet selector to explicitly select the duration and location within the video of the next segment that needs to be captioned. We first concatenate the whole video features $\mathcal{V}$, explicit commonsense feature $\mathbf{m}_{i-1}$, implicit commonsense feature $\mathbf{g}_{i-1}$, and hidden feature $\mathbf{h}_{i-1}$ from the last captioned snippet.
% as $V^*$. 
% Then we use $V^*$ as the 
These form the input to the snippet selector $f_n$ for which we leverage $3$ feed-forward layers with the \texttt{Sigmoid} activation function. The outputs $\mathbf{n}_{i} \in \mathbb{R}^{T}$ will be the probabilities over the video frames predicting the selected frames.

\begin{equation}
    \mathbf{n}_{i} = f_n(\text{concat}(\mathcal{V}, \mathbf{m}_{i-1}, \mathbf{g}_{i-1}, \mathbf{h}_{i-1})),
\end{equation}
This module can be trained with a binary-cross entropy loss objective, ${\mathcal{L}_{snippet}}$ as follows:
\begin{equation}\label{loss_snippet}
    \mathcal{L}_{snippet} = L_{BCE}(\mathbf{n}_i, \mathbf{n}_{GT}),
\end{equation}
where $\mathbf{n}_{GT}$ denotes the ground truth snippet duration using per frame binary (0/1) labels. 

\vspace{0.07in}
\noindent{\textbf{Action-Object Predictor.}}\label{subsubsec:actobj}
We observe that actions presented and objects appearing in the video are highly related to the corresponding sentence. Here, we introduce an action-object predictor module $f_a$ to predict the actions that happened and the objects that appeared in the snippet.
For certain datasets, such labels maybe available, \eg, the ALFRED
dataset \cite{shridhar2020alfred} contains $12$ labeled actions and $61$ objects. However, more generally, not all objects may be labeled, and not all datasets may contain this information. Hence, we leverage the pretrained CLIP \cite{radford2021learning} model to generate addition pseudo-labels for objects in a zero-shot manner (see \cite{radford2021learning} for details). In practice, we take each frame as the input to the pretrained CLIP and select the top $3$ predictions as pseudo-labels. Then, we concatenate both ground truth and pseudo labels for the snippet by considering objects or actions as present if they appeared in any of the frames belonging to the ground truth segment being considered. This results in a $12 + 61 + 604 = 677$-dimensional target for action-object predictions generated by this module. 

To train the action-object predictor $f_a$, we pull the video features from the predicted snippet segment $\mathbf{n}_{i}$ to get a snippet feature $\mathcal{V}_i$. We also concatenate to $\mathcal{V}_i$ commonsense feature $\mathbf{m}_{i-1}$, completion feature $\mathbf{g}_{i-1}$, and hidden feature $\mathbf{h}_{i-1}$ to generate a feature vector for action-object predictor. The outputs $\mathbf{a}_{i}$ is the probability distributions over all actions and objects. In detail, we formulate the action-object predictor $f_a$ with $3$ feed-forward layers with the \texttt{Sigmoid} activation function.
\begin{equation}
    \mathcal{V}_i = \mathcal{V} \otimes \mathbf{n}_i,
\end{equation}

\begin{equation}
    \mathbf{a}_{i} = f_a(concat(\mathcal{V}_i, \mathbf{m}_{i-1}, \mathbf{g}_{i-1}, \mathbf{h}_{i-1})),
\end{equation}
This module can be trained with a binary-cross entropy loss objective, ${\mathcal{L}_{actobj}}$ as follows:
\begin{equation}\label{loss_actobj}
    \mathcal{L}_{actobj} = L_{BCE}(\mathbf{a}_i, \mathbf{a}_{GT}),
\end{equation}
where $\mathbf{a}_{GT}$ denotes the concatenation of ground truth actions and objects (along with pseudo-labels from CLIP).

\begin{figure}[!t]
    \centering
    \includegraphics[width=0.5\textwidth]{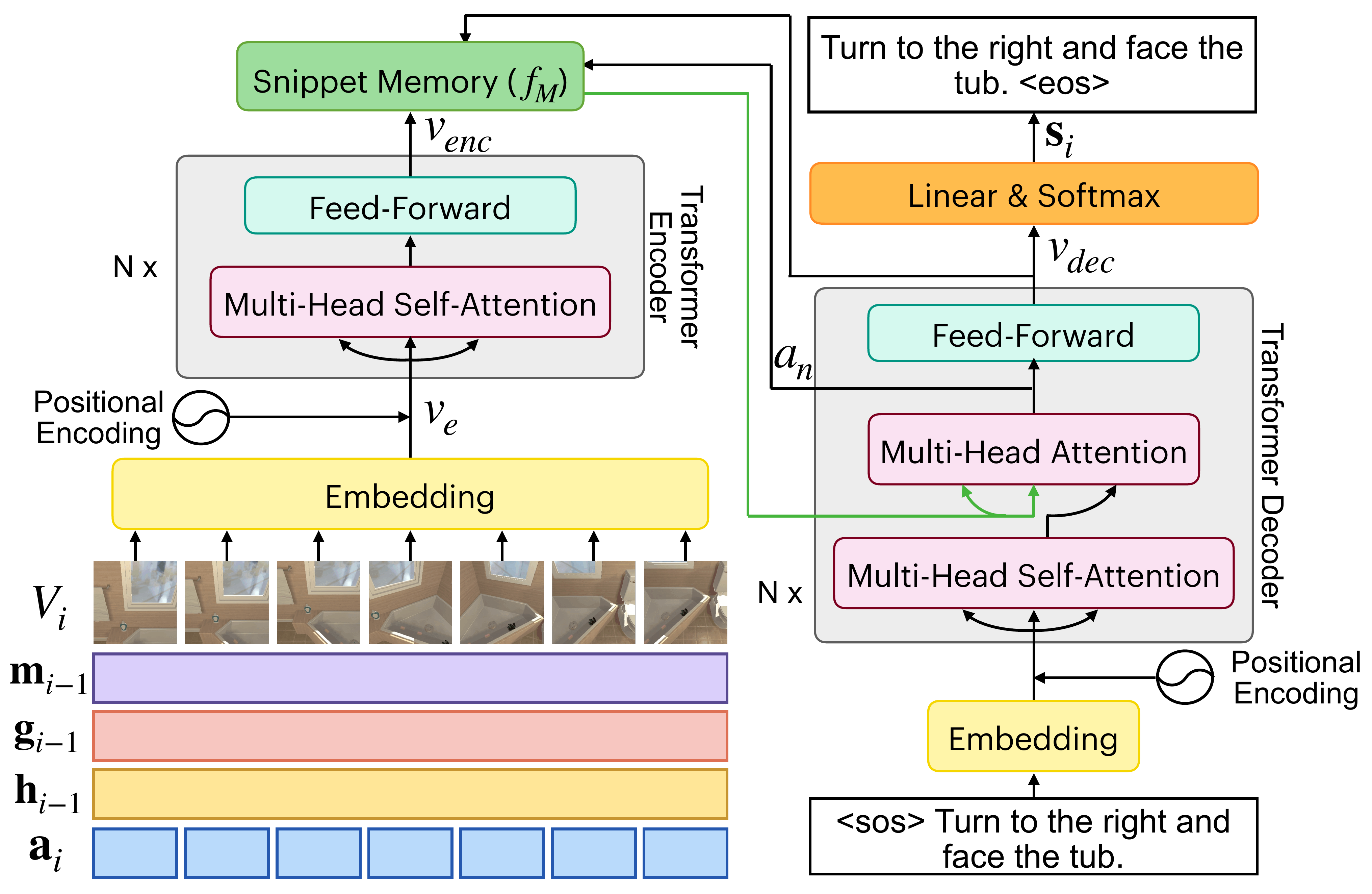}
    \caption{\textbf{Details of the Transformer-based Captioning Model.} The Transformer-based captioning model takes the snippet features, commonsense knowledge features, hidden features, and actions-objects predictions as the inputs of the Transformer encoder. The Snippet Memory leverages the ''add`` and ''erase`` operations to dynamically update the visual attention for the Transformer decoder at each decoding step. In the end, a Linear layer is used to output the generated sentence.}
    \label{fig:transformer}
\end{figure}

\subsection{Captioning Model}
We experiment with two types of Captioning Models: Transformer-based captioning model and LLM-based Foundational Captioning Model.

\vspace{0.07in}
\noindent{\textbf{Transformer-based captioning model}.}\label{subsubsec:transformer}
We use a video paragraph Transformer model~\cite{song2021towards} as the basis of the Transformer-based captioning model which consists of a Transformer encoder with the visual memory module and a Transformer decoder. Figure~\ref{fig:transformer} shows the architecture of the proposed caption model. The inputs of the Transformer encoder for snippet $i$ are the concatenation of visual feature $\mathcal{V}_i$, action feature $\mathbf{a}_i$, and the commonsense knowledge features obtained from the previous snippet for this one, $\mathbf{m}_{i-1}$ and $\mathbf{g}_{i-1}$, as well as the hidden features $\mathbf{h}_{i-1}$. After an embedding layer and positional encoding, $\mathcal{V}_e$ is derived and fed into the Transformer encoder.
In the $n$-th layer, the encoded feature $\mathcal{V}^{(n)}_e$ is computed as follows:
\begin{equation}
    \mathcal{V}^{(n)}_e = \mathcal{V}^{(n-1)}_e + MultiHead(\mathcal{V}^{(n-1)}_e, \mathcal{V}^{(n-1)}_e, \mathcal{V}^{(n-1)}_e).
\end{equation}
With the encoded feature, $\mathcal{V}_{e}$, we leverage the video memory from~\cite{song2021towards} to be our snippet memory module to enhance the temporal attention in the decoder. Instead of using the same $\mathcal{V}_{e}$ at each decoding step, the snippet memory $f_M$ dynamically updates the visual attention. $f_M$ has two operations, ``add'' and ``erase''. The goal of the ``add" operation is to progressively add more video clip features to the memory. On the other hand, the ``erase" operation is to weaken the already accessed features and encourages the model to describe more unseen parts. 
The output of the Snippet Memory module is then used as the key and value for the Multi-Head Attention layer in Transformer decoder.

After generating a sentence, $s_i$, using the decoder, we apply the typical MLE loss to enforce the model fits the ground-truth distribution. However, since the MLE loss tends to make the decoder output high-frequency tokens, we add high-frequency word penalties~\cite{welleck2020neural} to the objective function. Therefore, $\mathcal{L}_{sen}$ can be formulated as follows:
\begin{equation}\label{loss_sen}
\small
    \mathcal{L}_{sen} = -\frac{1}{J}\sum_{j=1}^{J}(\text{log}p(\mathbf{s}_i^j|\mathbf{s}_i^{<j}, \mathcal{V})) + \sum_{\mathbf{s}_i^* \in \mathbf{s}_i}\text{log}(1-p(\mathbf{s}_i^*|\mathbf{s}_i^{<j}, \mathcal{V})),
\end{equation}
where $J$ is the number of words in the sentence in the $i$-th snippet. $(\mathcal{V},\mathbf{s}_i)$ denotes the current training pair, $\mathbf{s}_i=\{\mathbf{s}_i^1,...,\mathbf{s}_i^{j-1}\} \setminus \{\mathbf{s}_i^j\}$ is the candidate word to be penalized. Therefore, we are able to not only emphasize the correct words but also decrease probabilities for the model to predict wrong candidates with high-frequency appearance.

\begin{figure}[!t]
    \centering
    \includegraphics[width=0.5\textwidth]{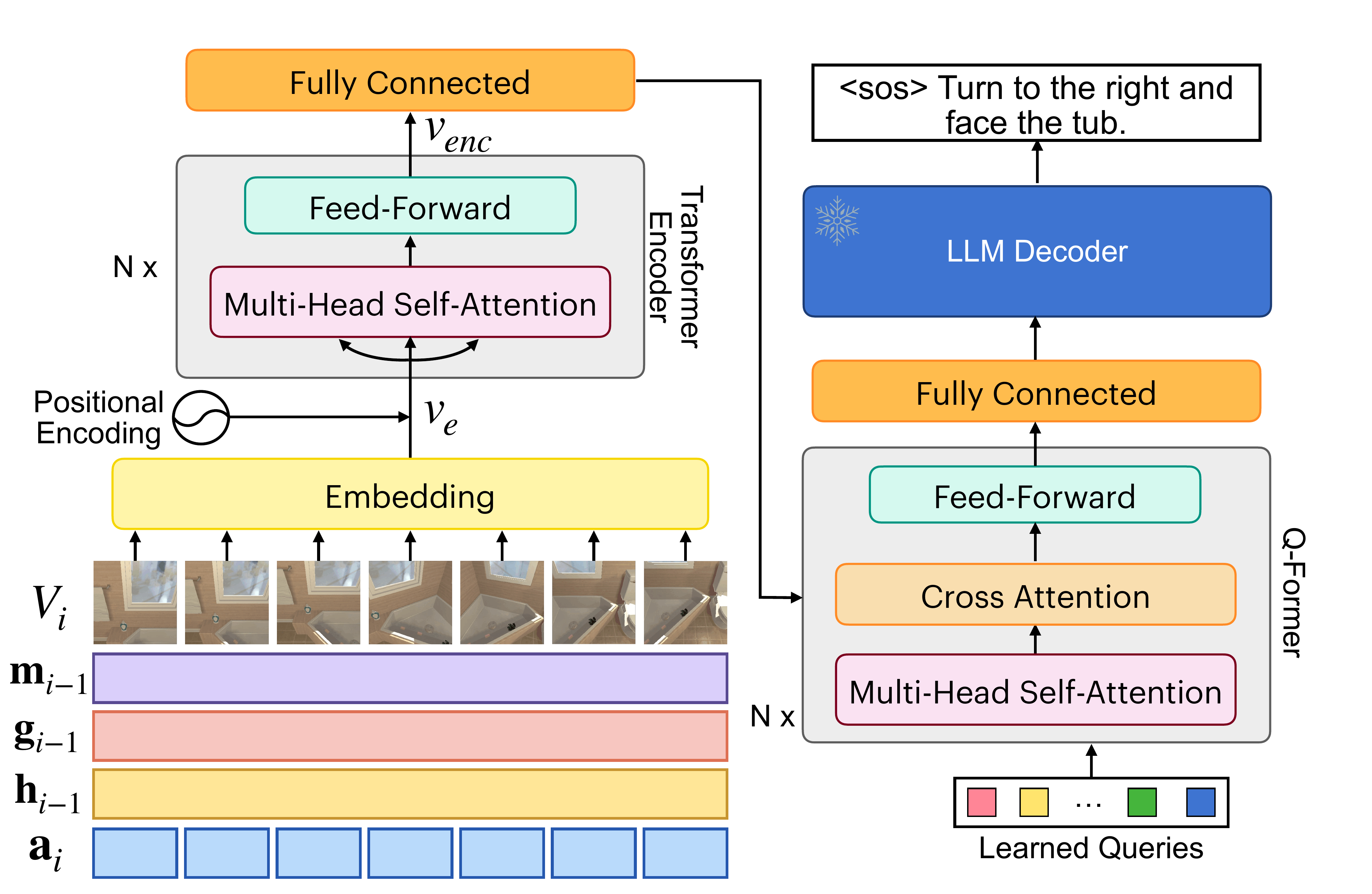}
    \caption{\textbf{Details of the LLM-based captioning model.}
    The LLM-based captioning model comprises the Transformer encoder, a Q-former, and a frozen LLM decoder. The transformer encoder takes the visual feature, action feature, commonsense knowledge features, and hidden feature as inputs, and a fully connected layer is applied to linearly project the features to the same dimension as the Q-former. The Q-former is a trainable module to bridge visual and language parts. In the end, the instructions are generated by the frozen LLM decoder.}
    \label{fig:LLM}
\end{figure}

\vspace{0.07in}
\noindent{\textbf{LLM-based Captioning Model.}}
With the rapid growth of the Large Language Model (LLM) and Multi-model LLM (MLLM), we leverage BLIP-2~\cite{li2023blip} as the basis of the LLM-based captioning model. The LLM-based captioning model consists of a visual Transformer encoder, a Q-former, and a LLM decoder. Figure~\ref{fig:LLM} shows the details of the MLLM-based captioning model. Similar to the Transformer-based captioning model, we concatenate the visual feature $\mathcal{V}_i$, the action feature $\mathbf{a}_i$, and the commonsense knowledge features obtained from the previous snippet, $\mathbf{m}_{i-1}$ and $\mathbf{g}_{i-1}$, as well as the hidden features $\mathbf{h}_{i-1}$ as the inputs to the Transformer encoder. A fully-connected layer is then applied to linearly project the encoded visual features into the same dimension as the pre-trained Q-former and LLM decoder. 
Q-Former is the trainable module to bridge the gap between the visual encoder and the frozen LLM. It consists of two transformer submodules that share the same self-attention layers. We initialize the Q-former with pre-trained parameters from BLIP-2 and fine-tune after. The pre-trained Q-former can effectively extract language-informative visual representation and feed the most useful information to the following LLM decoder while removing the irrelevance. The following frozen LLM is tasked to generate the text conditioned on the visual representation from Q-Former. For frozen LLM, we use the pre-trained LLM from BLIP-2.

\vspace{0.07in}
\noindent{\textbf{Hidden Feature.}}
To model the context between the generated captions, we use the pretrained SBERT to encode the generated captions to hidden features $\mathbf{h}_c \in \mathbb{R}^{384}$.

\subsection{Loss Function}\label{subsec:loss}
The overall training loss is a combination of the losses from Eq.~(\ref{loss_snippet}), Eq.~(\ref{loss_actobj}), and Eq.~(\ref{loss_sen}):
\begin{equation}\label{loss_overall}
\mathcal{L}= \lambda_{c}\mathcal{L}_{snippet} + \lambda_{a}\mathcal{L}_{actobj} + \lambda_{s}\mathcal{L}_{sen},
\end{equation}
where $\lambda_{c}$, $\lambda_{a}$, and $\lambda_{s}$ are hyper-parameters that modulate the relative importance of different terms.

\section{Instruction Generation}
The goal of the video-based instruction generation task, which we introduce, shares some similarity with dense and paragraph captioning, but also has important distinctions. 
The instruction generation task is defined as follows: given the first-person video of an actor performing a task, the model is required to generate the series of instructional sentences that outline steps in the task; this includes precise progression of actions and interactions with objects. 
While the inputs and outputs are similar, this problem is different from dense video captioning~\cite{krishna2017dense} in two key ways. 
First, the instructions require an understanding of allocentric and egocentric spatial expressions ({\em go right} and face the upper cabinet {\em to the left of the fridge}), verbs ({\em pick up} the pan from the stove), temporal guidance (turn on the microwave {\em for a few seconds}, {\em then} open the door and remove the heated apple). 
Second, commonsense knowledge is required in the context. Consider instruction: {\em turn around and go to the stove, then look up the microwave. open the microwave, put the egg inside, and close it.}; 
generating such a caption would be much easier if the agent knows that in order to use the microwave one needs to look at it first and opening of a microwave happens before closing. 
In contrast, in dense video captioning, which collects sparse data from the web, events are often independent and lack sufficient temporal and logical structure, making commonsense reasoning less relevant.

To construct a dataset for instruction generation  we leverage ALFRED dataset~\cite{shridhar2020alfred} which consists of natural language instructions and ego-centric vision for everyday tasks. We use \texttt{high\_idx}, annotated in the dataset, to split the videos into snippets. In total, there are $43,892$ snippets. The average frames per snippet is $42.8$. For each frame, we take the annotated actions and objects to form the action-object labels for training the action-object predictor. Notably, comparing with the actions in dense video captioning, the actions describe in the ALFRED dataset are relatively different (running, play a game, throw a ball v.s. pick up an apple, turn right). For the instructions, we take one example from each trial in each task. In total, there are $43,892$ step-by-step instructions. The average sentence length per instruction is $13.41$ words.

\section{Experiments}
% \subsection{Datasets}
We perform experiments on both the Alfred and ActivityNet Captions datasets for the instruction generation and dense video captioning tasks respectively.  

\vspace{0.05in}
\noindent \textbf{ALFRED Dataset. } 
The ALFRED dataset~\cite{shridhar2020alfred} is a benchmark for connecting actions, behaviors, and objects with human language. It is built in interactive visual environments on AI2-THOR 2.0~\cite{kolve2017ai2}. We take the language directive from each task and each trial. In total, we have $6,574$ for the training set, $506$ for the validation set, and $971$ for the testing set. Both validation and testing sets are split into seen and unseen folds. Scenes in the seen folds are the subsets in the training set. Scenes in the unseen folds are distinct from the training set and each other.

\vspace{0.05in}
\noindent \textbf{ActivityNet Captions Dataset. } 
The ActivityNet Captions dataset~\cite{krishna2017dense} contains $10,009/4,917/5,044$ videos for train/val/test, respectively. Each video has a series of temporally annotated sentences and each sentence corresponds to a segment of the video which describes an occurring event. We follow the same settings in~\cite{seo2022end, krishna2017dense} using ground truth temporal proposals to evaluate the predicted captions.

\begin{table*}
\small
\begin{center}
\caption{\textbf{Qualitative Results.} We evaluate the proposed video Transformer-based captioning model on the ALFRED~\cite{shridhar2020alfred} dataset. Our models including the commonsense knowledge outperform the baseline. The ablations for including different losses and modules are also provided. B@4, M and C are BLEU@4, METEOR and CIDEr respectively. Best results in bold.} %\alex{be specific} 
%\vspace{-1em}
\label{tab:ablation}
%\vspace{-1em}
\begin{tabular}{c|l|c c c c|c c c|c c c}
    \toprule
    & \multicolumn{1}{l|}{} & \multicolumn{1}{l}{} & \multicolumn{1}{l}{} & \multicolumn{1}{l}{} & \multicolumn{1}{l|}{} & \multicolumn{3}{c|}{\textbf{Seen}} & \multicolumn{3}{c}{\textbf{Unseen}} \\
    & Settings & $\lambda_{a}$ & $\lambda_{c}$ & COMET & LM & B@4 & M & C & B@4 & M & C \\
    \midrule
    & Paragraph-Level~\cite{song2021towards}  & & & & & 13.87 & 15.00 & 17.94 & 12.05 & 11.53 & 16.93 \\
    & ~~~+ Action-Object (w/o CLIP) & \checkmark & & & & 14.32 & 13.87 & 17.80 & 13.51 & 10.66 & 17.39\\
    & ~~~+ Action-Object (w/ CLIP)  & \checkmark & & & & 14.38 & 16.47 & 18.17 & 13.86 & 12.61 & 17.15\\
    \midrule
    \parbox[t]{2mm}{\multirow{5}{*}{\rotatebox[origin=c]{90}{Transformer}}} & Sentence-Level & \checkmark & \checkmark & & & 15.38 & 17.84 & 18.13 & 13.32 & 12.58 & 17.36\\
    & ~~~+ LM  & \checkmark & \checkmark & & \checkmark & 15.70 & 19.03 & 18.38 & 14.34 & 16.38 & 17.79\\
    & ~~~+ COMET (1 relation type)  & \checkmark & \checkmark & \checkmark & & 16.27 & 21.78 & 18.37 & 14.35 & 17.23 & 17.58\\
    & ~~~+ COMET  & \checkmark & \checkmark & \checkmark & & 16.59 & \textbf{23.66} & 19.03 & 14.39 & \textbf{19.40} & 17.65\\
    & ~~~+ COMET \& LM  & \checkmark & \checkmark & \checkmark & \checkmark & \textbf{17.18} & 23.06 & \textbf{19.48} & \textbf{15.41} & 17.94 & \textbf{18.46}\\
    \midrule
    \parbox[t]{2mm}{\multirow{5}{*}{\rotatebox[origin=c]{90}{LLM}}} & Sentence-Level & \checkmark & \checkmark & & & 14.32 & 19.03 & 18.87 & 13.25 & 13.91 & 17.45 \\
    & ~~~+ LM  & \checkmark & \checkmark & & \checkmark & 14.33 & 19.42 & 19.04 & 13.76 & 16.89 & 17.82 \\
    & ~~~+ COMET (1 relation type)  & \checkmark & \checkmark & \checkmark & & 15.11 & 21.93 & 18.55 & 13.81 & 17.46 & 17.65 \\
    & ~~~+ COMET  & \checkmark & \checkmark & \checkmark & & 16.68 & 22.35 & 19.23 & 15.34 & 17.92 & 17.88\\
    & ~~~+ COMET \& LM  & \checkmark & \checkmark & \checkmark & \checkmark & \textbf{17.53} & \textbf{23.45} & \textbf{19.55} & \textbf{16.02} & \textbf{18.23} & \textbf{18.72} \\
    \bottomrule
\end{tabular}
\end{center}
\end{table*}

\vspace{0.05in}
\noindent \textbf{Evaluation Metrics. }
We evaluate metrics widely used for visual captioning tasks~\cite{krishna2017dense, lei2020mart, park2019adversarial, xiong2018move}. The generated captions are evaluated against the ground truth with three traditional metrics: BLEU~\cite{papineni2002bleu}, METEOR~\cite{denkowski2014meteor}, and CIDEr~\cite{vedantam2015cider}.

\vspace{0.05in}
\noindent {\bf Implementation Details.}
For both datasets, the videos are truncated to a maximum number of $150$ frames at FPS=$1$ (in other words 150 seconds or 2.5 minutes). %\leon{150 frames? at 30FPS? It must be lower FPS. Lets clarify.}
The maximum number of snippets is set to $20$.
To extract video features, we use ResNet-200~\cite{he2016deep} pretrained on ImageNet and I3D (RGB+Flow)~\cite{carreira2017quo} pretrained on the Kinetics dataset. 
We divide the videos into non-overlapping visual tokens with $12$ frames for the ALFRED and $64$ for the ActivityNet Captions dataset. 
For text, we truncate sentences to a maximum length of $150$ for both datasets.
The vocabulary dictionary is $3,260$ words for the ALFRED and $10,246$ words for the ActivityNet Captions dataset.
The Transformer encoder and decoder layers are both set to $N=3$. The hidden size is set to $512$; the number of attention heads is $8$. During training, we use label smoothing~\cite{szegedy2016rethinking} with the value set to $0.1$ and optimize with the learning rate varied under a warm-up strategy with $2,000$ steps for the ALFRED and $8,000$ steps for the ActivityNet Captions dataset. In the inference phase, we generate the paragraph with a greedy search. We train the model with batch size $4$ on the ALFRED and batch size $2$ on the ActivityNet Captions dataset on a single GPU.

\subsection{Results on Instruction Generation}
\noindent \textbf{Baselines.} 
We compare with the natural baseline~\cite{song2021towards} which uses the Transformer encoder-decoder with a video memory module to generate the paragraph from videos. %We also compare with an additional baseline, PDVC~\cite{wang2021end}. The results are in the supplementary.
%Since our model is a direct extension of the paragraph captioning model~\cite{song2021towards}, we use it as natural baseline.

\vspace{0.07in}
\noindent \textbf{Abaltion Study.}
We consider the impact components in the proposed model make on performance in Table~\ref{tab:ablation}. Specifically, we incrementally add one component at a time to the baseline model. This results in the following variants:
\begin{itemize}
\itemsep 0em 
\item[] [+ Action-Object (w/o CLIP)]: Adds Action-Object Predictor. For this we only consider actions and objects that are labeled in the training set.
\item[] [+ Action-Object (w/ CLIP)]: The same settings as {\em Action-Object (w/o CLIP)} but with additional objects and supervision from pseudo-labels given by CLIP.
\item[] [+ COMET (1 relation type)]: Adds explicit commonsense knowledge obtained using the pretrained COMET model. The `1 relation' variant only uses \texttt{<AtLocation>} relation.
\item[] [+ COMET]: Adds explicit commonsense knowledge obtained using the pretrained COMET model with the selected $12$ relation types. 
\item[] [+ LM]: Adds implicit commonsense knowledge obtained using the pretrained GPT-Neo model.
\end{itemize}
The hyper-parameters of $\lambda_{c}=10$, $\lambda_{a}=10$ and $\lambda_{s}=1$ in Eq.~(\ref{loss_overall}) are chosen through cross-validation. 

\begin{figure*}[!t]
    \centering
    \includegraphics[width=\textwidth]{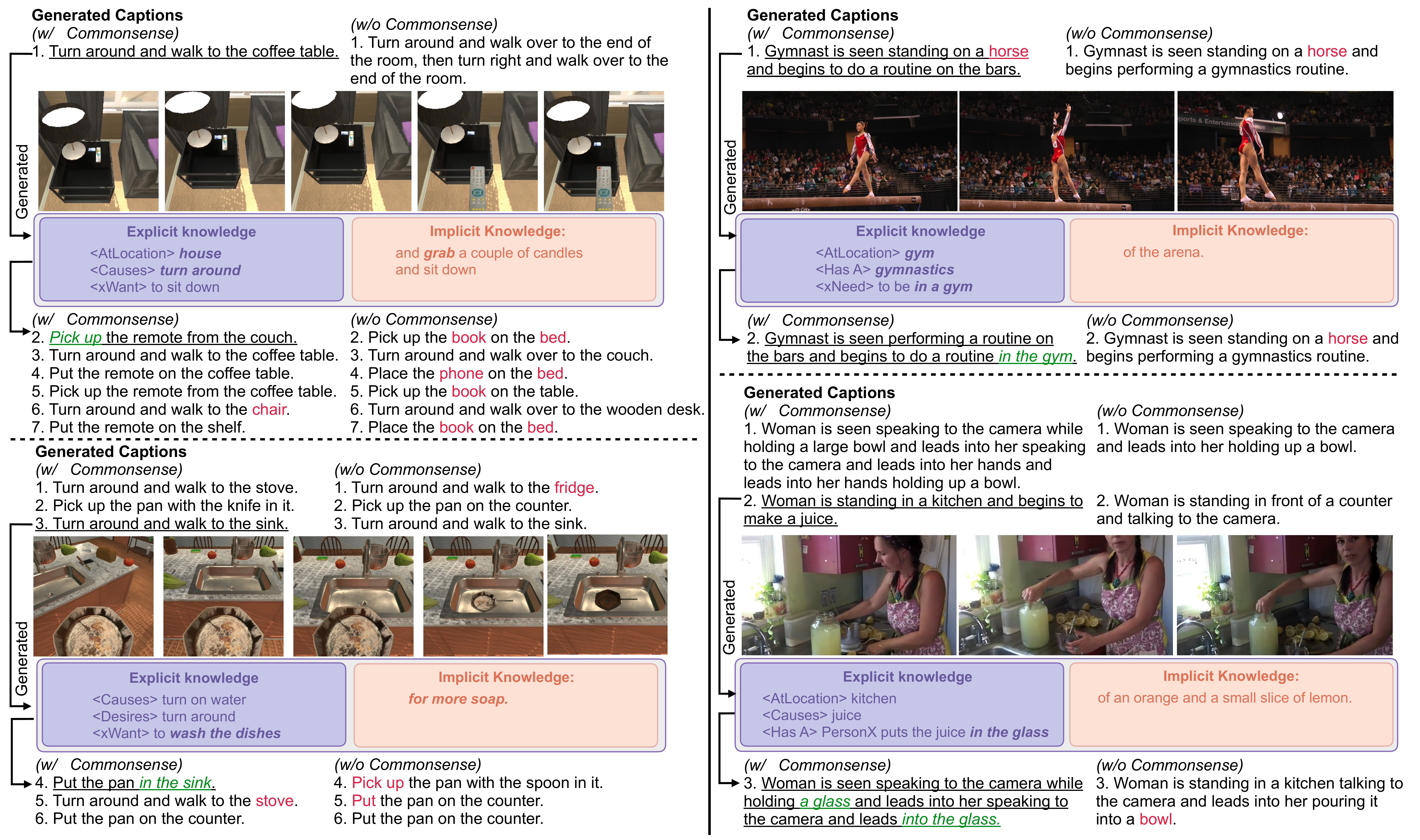}
    \caption{\textbf{Qualitative Results.} We show two qualitative examples from each dataset. Illustrated are generated captions obtained by our model; for comparison we also show results from a baseline variant without (w/o) commonsense knowledge. The generated explicit and implicit knowledge are also provided for one intermediate sentence. As shown in the figure, explicit knowledge provides relations like \texttt{AtLocation}, and \texttt{xNeed} that the model can and does leverage. Implicit knowledge includes \textit{grab}, and \textit{for more soup} that correspond to the following actions, \textit{pick up}, and \textit{rinse it off}, in our generated and ground truth captions. The red color indicates the wrong predictions. One can also see that commonsense in our model leads to less repetition across sentences, higher accuracy and, often, richer descriptions.}
    \label{fig:qualitative}
\end{figure*}

\begin{table}
\small
\begin{center}
\caption{\textbf{Results on ActivityNet Captions Dataset.} We compare baselines trained on ActivityNet Captions dataset only and the model with large-scale pretraining (the last row). B@n denotes BLEU@n. M is the METEOR score.}%\vspace{-1em}
\label{tab:all}
\begin{tabular}{l|c|c|c}
    \toprule
    Method & B@3 & B@4 & M \\
    \midrule
    DCEV~\cite{krishna2017dense} & 4.09 & 1.60 & 8.88 \\
    DVC~\cite{li2018jointly} & 4.51 & 1.71 & 9.31 \\
    Bi-SST~\cite{wang2018bidirectional} & - & - & 10.89 \\
    HACA~\cite{wang2018watch} & 5.76 & 2.71 & 11.16 \\
    MWSDEC~\cite{rahman2019watch} & 3.04 & 1.46 & 7.23 \\
    MDVC~\cite{iashin2020multi} & 5.83 & 2.86 & 11.72 \\
    BMT~\cite{iashin2020better} & 4.62 & 1.99 & 10.90 \\
    iPerceive DVC~\cite{Chadha2020iPerceive} & 6.13 & 2.98 & {\bf 12.27} \\
    \midrule
    Ours (no commonsense) & 6.25 & 4.57 & 10.86 \\
    Ours (Transformer-based) & {\bf 7.01} & {\bf 5.27} & 11.40 \\
    Ours (LLM-based) & {\bf 7.55} & {\bf 5.86} & 12.05 \\
    \midrule \midrule
    MV-GPT~\cite{seo2022end} & - & 6.84 & 12.31 \\
    \bottomrule
\end{tabular}
\label{tab:activity}
\end{center}
\end{table}

\vspace{0.07in}
We test performance in two settings: (i) {\em Paragraph-Level}: one where snippets are not estimated or supervised (\ie, left latent); this is the setting of our baseline \cite{song2021towards}.
(ii) {\em Sentence-Level}: one where we learn to estimate snippets and utilize them in captioning, similar to dense video captioning.
In detail, in the paragraph-level generation, we take the whole video as input and directly generate multi-sentence instructions.  
In the (our full model) sentence-level setting, the model generates one instruction at a time iteratively, so the snippet selector is needed to predict the temporal location of the video to look at. The model can leverage commonsense by querying with the previously generated instruction under this setting.

In the {\em Paragraph-Level} setting, compared with the baseline~\cite{song2021towards}, adding the predicted actions and objects helps the model better understand the videos and generate more relevant captions. As a result, we leverage this setting for all experiments in the Sentence-Level.
In the {\em Sentence-Level}, all experiments outperform the baseline. Comparatively, using explicit commonsense knowledge is better than implicit knowledge. However, we can also see a level of complementary. Further, multiple COMET relations are shown to be useful. 

\vspace{0.04in}
\noindent{\textbf{Quantitative and Qualitative Results.}} 
Our improvements in instruction generation over the \cite{song2021towards} baseline are substantial. 
Specifically, the improvement in METEOR score is $\sim 8$ points [$15.00 \rightarrow 23.06$] (or, relatively speaking, over 50\%) on Seen and similar on Unseen data fold. It shows a better improvement with the LLM-based captioning model. The {\em Seen} fold consists of the scenes that are in the train set; the {\em Unseen} fold consists of entirely new scenes unencountered in training. Improvements on B@4 are also sizable, \eg, [$13.87 \rightarrow 17.18$] on the Transformer-based caption model and [$13.87 \rightarrow 17.53$] on the LLM-based caption model. 
The ablations in Table~\ref{tab:ablation} reveal significant improvements when incorporating different commonsense. Notably, COMET demonstrates a greater contribution than LM; augmenting the number of relations from COMET further improves the performance.
In addition, we calculate the accuracy of predicted actions and snippets, which are $73.72\%$ and $80.33\%$, respectively.
We show two qualitative results from Alfred and ActivityNet datasets in Fig~\ref{fig:qualitative}. The generated captions and the corresponding {\em explicit} and {\em implicit} knowledge are illustrated for an intermediate sentence in each video. Explicit knowledge provides information such as \texttt{<xWant>} \textit{wash the dishes} or \texttt{<AtLocation>} \textit{gym} that are related to the next snippet.
Implicit knowledge such as \textit{grab} and \textit{for more soap} is related to the next generated caption which illustrates its effectiveness.

\subsection{Results on Dense Video Captioning}
To illustrate generality, we also demonstrate the performance of our method on the ActivityNet Captions dataset~\cite{krishna2017dense} in the dense video captioning problem setting. 
We compare our results to prior works published for this task in Table~\ref{tab:activity}. 
Our model with commonsense knowledge outperforms the baselines (that, like us, only use ActivityNet data), setting SoTA on B@3 and B@4 scores. Compared with iPerceive DVC \cite{Chadha2020iPerceive}, which is the only approach that leverages commonsense knowledge, our model improves by a margin (especially on B@4). 
Our performance using METEOR is also competitive.
Further, we compare favorably with MV-GPT~\cite{seo2022end}. Unlike other methods, it is pretrained with a large-scale dataset, HowTo100M dataset~\cite{miech2019howto100m}, before finetuning with the ActivityNet Captions dataset. 

% \vspace{-.5em}
\subsection{User Study}
% \vspace{-.5em}
We conducted a user study to quantify our improvement with commonsense.
We select $23$ clips from $4$ examples for instruction generation and $11$ clips from $5$ examples for dense video captioning. The participants are asked to indicate whether sentence A is better, sentence B is better, both are better, or both are bad. In the end, we collect $10$ replies in the instruction generation and $12$ in dense video captioning. Human judges decide that $82.6\%$ of sentences from the model with adding commonsense are on par or better than those from the model without adding commonsense in instruction generation and $54.5\%$ in dense video captioning.

%\vspace{-1em}
\section{Conclusion}
We propose a new task of instruction generation which is motivated by imitation learning. It requires the generation of interrelated sentences. Furthermore, we propose a novel video captioning Transformer-based model that takes implicit and explicit commonsense knowledge into account. This formulation enhances the quality of the generated captions and makes the generated instructions equip the abilities such as persistence, spatial awareness, and functional inference. Experiments on the ALFRED dataset and the ActivityNet Captions dataset demonstrate improvements in the BLEU, METEOR, and CIDEr metrics. 

\section*{Acknowledgement}
This work was funded, in part, by the Vector Institute for AI, Canada CIFAR AI Chair, NSERC CRC, and NSERC DG grants. Hardware resources used in preparing this research were provided, in part, by the Province of Ontario, the Government of Canada through CIFAR, and companies sponsoring the Vector Institute\footnote{www.vectorinstitute.ai/partners}. Additional hardware support was provided by John R. Evans Leaders Fund CFI grant and Compute Canada under the Resource Allocation Competition award.

{\small
\bibliographystyle{ieee_fullname}
\bibliography{egbib}
}

\end{document}